\documentclass[11pt,a4paper]{article}
\pdfoutput=1

\usepackage{times,latexsym}
\usepackage{url}
\usepackage[T1]{fontenc}

\usepackage[acceptedWithA]{tacl2018v2}

\usepackage[]{tacl2018v2}

\usepackage{xspace,mfirstuc,tabulary}

\newif\iftaclinstructions
\taclinstructionsfalse %
\iftaclinstructions
\renewcommand{\confidential}{}
\renewcommand{\anonsubtext}{(No author info supplied here, for consistency with
TACL-submission anonymization requirements)}
\newcommand{\instr}
\fi

\iftaclpubformat %

\else

\fi

\usepackage{tikz}
\usepackage{pgfplots}
\usepgfplotslibrary{fillbetween}
\usepgfplotslibrary{groupplots}
\usepackage{amsmath}
\usepackage{soul}
\usepackage{bbm}
\usepackage{setspace}
\usepackage{multirow}
\usepackage{bigdelim}
\usepackage{amsfonts}
\usepackage{balance}
\usepackage[T1]{fontenc}
\usepackage{amsmath,amsfonts,amssymb,amsthm}
\usepackage{dsfont}
\usepackage{xspace}
\usepackage{ulem}
\usepackage{adjustbox}

\usepackage{enumitem}

\usepackage{booktabs}

\definecolor{plotcolor1}{RGB}{35,127,215}
\definecolor{plotcolor2}{RGB}{46,159,52}
\definecolor{plotcolor3}{RGB}{227,118,18}
\definecolor{plotcolor4}{RGB}{142, 68, 173}
\definecolor{plotcolor5}{RGB}{255,51,51}
\definecolor{plotcolor6}{RGB}{241, 196, 15}
\definecolor{plotcolor7}{RGB}{100,118,18}
\definecolor{plotcolor8}{RGB}{33, 47, 61}
\definecolor{plotcolor9}{RGB}{131, 145, 146}
\definecolor{plotcolor10}{RGB}{211, 84, 0}

\newcommand*{\Courier}{\fontfamily{pcr}\selectfont}
\DeclareTextFontCommand{\textCourier}{\Courier}

\usepackage{tikz}
\usepackage{pgfplots}
\usetikzlibrary{matrix}
\usepgfplotslibrary{groupplots}
\usetikzlibrary{fit}
\pgfplotsset{compat=newest}
\usepgfplotslibrary{colorbrewer}

\usepackage{pgfplotstable}
\pgfplotstableset{col sep=comma}

\usepackage{balance}
\usepackage{xspace}

\usepackage{graphicx}
\usepackage{caption}
\usepackage{subcaption}

\usepackage{titlesec}
\titlespacing{\paragraph} {0pt}{0.1ex plus 0.1ex minus .1ex}{1em}

\makeatletter
\g@addto@macro\small{%
  \setlength\abovedisplayskip{-5pt}
  \setlength\abovedisplayshortskip{-5pt}
  \setlength\belowdisplayshortskip{-5pt}
  \setlength\belowdisplayskip{-5pt}
}
\makeatother

\usepackage{algorithmicx} 
\usepackage[noend]{algpseudocode}
\usepackage{algorithm}
\usepackage[T1]{fontenc}

\algnewcommand\algorithmicdefinitions{\textbf{Definitions:}}
\algnewcommand\Definitions{\item[\algorithmicdefinitions]}
\renewcommand{\algorithmiccomment}[1]{{\color{gray}\raisebox{1px}{\texttt{\guillemotright}} #1}}
\algnewcommand{\LineComment}[1]{\Statex \hskip\ALG@thistlm \algorithmiccomment{#1}}
\algrenewcommand\alglinenumber[1]{\footnotesize #1:}
\algrenewcommand\algorithmicindent{1.0em}%
\makeatletter
\newcommand{\StatexIndent}[1][3]{%
  \setlength\@tempdima{\algorithmicindent}%
  \Statex\hskip\dimexpr#1\@tempdima\relax}

\newcommand{\nlstring}[1]{\textit{#1}}
\newcommand{\reals}{{\rm I\!R}}

\newcommand{\system}[1]{\textsc{#1}}

\newcommand{\cerealbar}{\textsc{CerealBar}\xspace}

\newcommand{\sentence}{\bar{x}}

\newcommand{\token}{x}

\newcommand{\state}{s}

\newcommand{\startstate}{\state_1}

\newcommand{\pathplan}{\bar{p}}

\newcommand{\pathplanlabel}{\bar{\rho}}

\newcommand{\round}{r}

\newcommand{\dataset}{\mathcal{D}}

\newcommand{\heightidx}{h}
\newcommand{\widthidx}{w}
\newcommand{\rotidx}{\alpha}

\newcommand{\feedback}{f}

\newcommand{\sentlabel}{y}

\newcommand{\ipscoeff}{\ell}

\newcommand{\pose}{p}
\newcommand{\poserep}{\mathbf{p}}

\newcommand{\modelprob}{P}

\newcommand{\execution}{\bar{e}}

\newcommand{\embed}{\phi}
\newcommand{\embedstate}{\phi^\state}

\newcommand{\conv}{\textrm{CNN}}

\newcommand{\params}{\theta}

\newcommand{\sysfull}{\system{Full}\xspace}
\newcommand{\sysposonly}{\system{Pos-Only}\xspace}
\newcommand{\systconly}{\system{TC-Only}\xspace}
\newcommand{\sysnoensemble}{\system{No-Ensemble}\xspace}
\newcommand{\sysfinetune}{\system{Fine-Tuning}\xspace}

\newcommand{\soutthick}[1]{%
    \renewcommand{\ULthickness}{1.2pt}%
       \sout{#1}%
    \renewcommand{\ULthickness}{.4pt}%
}
\newcommand{\erase}[1]{\textcolor{red}{\soutthick{\textcolor{black}{#1}}}}

\newcommand{\textcorrection}[1]{\underline{\textcolor{blue}{#1}}}
\newcommand{\textred}{\textcolor{red}}

\title{Continual Learning for Grounded Instruction Generation \\ by Observing Human Following Behavior}

\author{Noriyuki Kojima, Alane Suhr, \textnormal{and} Yoav Artzi  \\
Department of Computer Science and Cornell Tech, Cornell University\\ 
{\tt nk654@cornell.edu} \hspace{1em}
{\tt \{suhr,  yoav\}@cs.cornell.edu}\\ }

\date{}

\begin{document}

\maketitle

\begin{abstract}
We study continual learning for natural language instruction generation, by observing human users' instruction execution. We focus on a collaborative scenario, where the system both acts and delegates tasks to human users using natural language. We compare user execution of generated instructions to the original system intent as an indication to the system's success communicating its intent. We show how to use this signal to improve the system's ability to generate instructions via contextual bandit learning. In interaction with real users, our system demonstrates dramatic improvements in its ability to generate language over time.

\end{abstract}

\section{Introduction}\label{sec:intro}

Natural language provides an expressive and accessible avenue to instruct non-expert users. 
The ability to generate instructions is critical for systems that collaborate with users, for example to delegate tasks. 
In such scenarios, the system generates language to communicate to the user a latent intent. 
When users are cooperative and proficient in the language, whether they accomplish the system's intent provides an informative, albeit noisy signal to the quality of instruction generation. 

This implicit signal is fundamentally different from supervised data, including via active learning, in that it does not label the system's intent with a written instruction, but only provides evidence to the quality of a given instruction in relaying this intent. 
As a natural byproduct of interaction with users, it also differs from explicit user feedback in not requiring user action beyond what they already do as part of the interaction. 
Despite its potential and prevalence, this signal is understudied for learning to generate natural language.

\begin{figure}
    \centering
    \footnotesize
    \includegraphics[width=\linewidth,clip,trim=75 123 32 40]{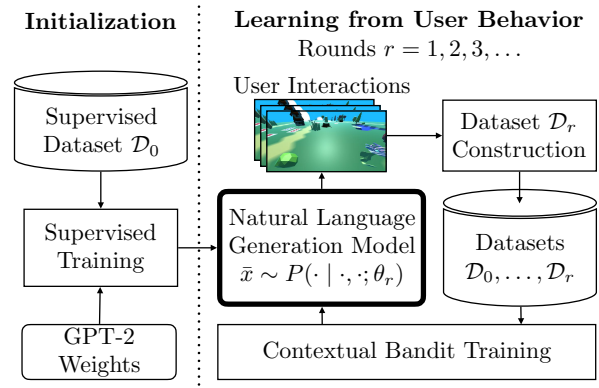}
    \caption{Diagram of our learning process. We initialize a generation model using supervised learning, and continually learn through interaction with users, by alternating between observing user execution of generated instructions and training.}
    \label{fig:intro}
    \vspace{-5pt}
\end{figure}

In this paper, we study this learning signal. 
We formalize continually improving instruction generation by observing human users executing generated instructions.
We learn by  comparing instruction execution to the system intent, and demonstrate how this results in a system that continually improves its natural language generation ability through interaction with users. Figure~\ref{fig:intro} illustrates our learning process.

We design a task-oriented collaborative scenario using the \cerealbar game environment~\cite{Suhr2019:cerealbar}. 
In \cerealbar, two agents, a leader and a follower, work together to complete tasks. 
The leader plans the tasks to complete, and communicates goals to the follower using natural language. 
\cerealbar was originally introduced for studying follower instruction execution. We modify it to focus on generation of leader instructions, which are then executed by human followers. The collaborative, embodied setup effectively engages users, and aligns their incentives with executing the system's instructions to the best of their abilities.

A major challenge is inferring a learning signal from observed user behavior. 
Given the user execution, we create positive and negative examples, depending on how the user execution aligns with the system's plan and the user's perceived correctness of their own execution. 
For example, consider an execution that does not align well with the system's plan, but that the user considers correct given the instruction. 
Because of the misalignment, we cannot consider the instruction as a successful example given the system's plan. 
However, given the user's perceived correctness, we can generate a positive example treating the user's execution as a plan paired with the instruction. 
In contrast to supervised learning with gold-standard per-token labels~\cite{Sutskever:14}, such utterance-level binary labels form a challenging signal for learning, because they do not distinguish between correct and incorrect tokens. 

We do not make the typical distinction between training and deployment; as human users follow generated instructions, we continually collect new data, periodically train using this data, and evaluate the system through the interaction itself. 
We formalize learning as an off-policy contextual bandit learning problem. 
We show that positive examples can be treated in a manner that reduces to supervised learning, allowing for simple effective use of the data. 
However, using negative examples is more challenging, because simply minimizing their likelihood gives an unbounded negative loss. 
We weigh negative examples using an inverse propensity score~\cite[IPS;][]{Horvitz1952:ips,Wang2017:offpolicy-cb} to address this issue.

We experiment with our approach through interaction with human users, tracking both task performance and how the generated language changes. 
We observe dramatic improvements in the quality of instructions generated as reflected in users' execution: task completion in accordance to the system intent increases from 44.7\% to 79.3\%. 
This is accompanied by significant language change: the occurrence of erroneous phrases decreases as desired, but the effective system vocabulary gradually shrinks.

Although using user feedback for improving language generation has been studied, as we discuss in Section~\ref{sec:related}, to the best of our knowledge, this study is the first to show effective instruction generation learning by observing user execution. 
Our experiments demonstrate the effectiveness of our process, but also illustrate limitations and important directions for future work. Code and data are available at \href{https://lil.nlp.cornell.edu/cerealbar/}{https://lil.nlp.cornell.edu/cerealbar/}.

\section{Technical Overview and Notation}\label{sec:overview}

Our goal is  to continually improve a natural language instruction generation model, by observing human executions of generated instructions.

\paragraph{Interaction Scenario}
We focus on a collaborative scenario, where two agents, a leader and a follower, complete tasks in an environment. 
The system is the leader, and the human user is the follower. 
The leader plans tasks to accomplish, acts in the world, and instructs the follower using natural language. 
We use a deterministic procedure for planning and executing leader actions, and focus on learning the leader instruction generation model. 
The human follower acts in the world following the system instructions. 
We instantiate this scenario using \cerealbar (Section~\ref{sec:scenario}), a collaborative game, where two agents collect sets of cards together by moving in a 3D environment. 

\paragraph{Task}
A  world state $\state$ describes the current environment; in \cerealbar, this includes the location of landmarks, cards, and both agents. 
A plan $\pathplan$ is a sequence of poses $\langle  \pose_1,\dots,\pose_{|\pathplan|}\rangle$ the system intends for the human user to take starting from a start state $\startstate$. 
In \cerealbar, a plan includes moving in the environment with the intent of collecting cards; each pose $\pose_j$ is a tuple $(\heightidx_j, \widthidx_j, \rotidx_j)$, where  $\heightidx_j$ and $\widthidx_j$ are height and width coordinates, and $\rotidx_j$ is a discrete orientation angle. 
An instruction $\sentence$ is a sequence of tokens $\langle \token_1, \dots,\token_{|\sentence|} \rangle$. 
An instruction execution $\execution$ is the sequence of poses $\langle \pose_1, \dots, \pose_{|\execution|} \rangle$ a user takes executing $\sentence$, starting in a start state $\startstate$.
The generation distribution $\modelprob(\sentence \mid \startstate, \pathplan; \params)$ is parameterized by $\params$. 
The goal of instruction generation is that given a generated instruction $\sentence \sim \modelprob(\cdot \mid \startstate, \pathplan; \params)$, the user execution $\execution$ from $\startstate$ will follow the plan $\pathplan$. 
The user does not have access to $\pathplan$, but only to its description $\sentence$. 

\paragraph{Learning}

We use an encoder-decoder neural network model (Section~\ref{sec:model}), which we continually improve by observing user behavior.
This process proceeds in rounds. 
At each round $\round$, we first collect data and then train our model by estimating the model parameters $\params_\round$. 
During data collection in round $\round$, we sample from our model to generate instructions, and observe a human user's execution of each instruction. 
An execution of an instruction $\sentence \sim \modelprob(\cdot \mid \startstate, \pathplan ; \params_\round)$ generated for the plan $\pathplan$ with start state $\startstate$ creates a tuple $(\startstate, \pathplan, \sentence, \execution, \feedback)$, where $\execution$ is the user execution and $\feedback$ is structured user feedback solicited using binary questions (e.g., about the grammaticality of $\sentence$). 
The learner does not observe the user's actions executing $\sentence$, but only their poses along the execution. 
Given these tuples, we create a dataset $\dataset_\round = \{(\startstate^{(i)}, \pathplanlabel^{(i)}, \sentence^{(i)}, \sentlabel^{(i)})\}_{i=1}^{|\dataset_\round|}$, where $\sentlabel^{(i)}\in\{-1,+1\}$ is a binary label. 
Depending on the user execution and feedback, the plan $\pathplanlabel^{(i)}$ is either the original plan $\pathplan^{(i)}$ used for generating $\sentence^{(i)}$ or the user execution $\execution^{(i)}$ of $\sentence^{(i)}$. 
We formulate estimating $\params_{\round+1}$ as a contextual bandit learning problem with $\sentlabel$ as the reward.  
Section~\ref{sec:learn} describes the complete learning process. 

\paragraph{Evaluation}
Throughout the system's lifetime, we measure how well human users complete tasks, and also use earth mover's distance~\cite[EMD;][]{Rubner1998:emd} to quantify the similarity of the user execution $\execution$ to the plan $\pathplan$. 
We characterize language change over time by tracking vocabulary size, instruction length, and other statistics.

\section{Interaction Scenario}\label{sec:scenario}

\begin{figure}
    \centering
    \footnotesize
    \includegraphics[width=\linewidth,clip,trim={122 268 132 97}]{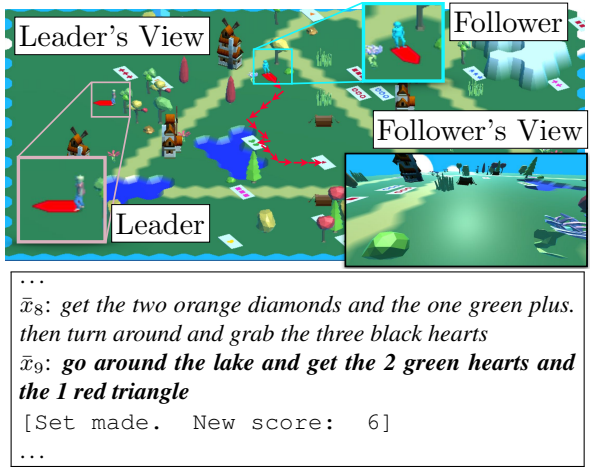}
    \fbox{\begin{minipage}[]{0.95\linewidth}
        \dots \\
        $\sentence_8$: \nlstring{get the two orange diamonds and the one green plus. then turn around and grab the three black hearts} \\
        $\sentence_9$: \textbf{\nlstring{go around the lake and get the 2 green hearts and the 1 red triangle}}\\
        \texttt{[Set made. New score: 6]} \\
        \dots
    \end{minipage}}
    \caption{Interaction snapshot in \cerealbar, with instructions generated by our model.
    The current instruction is $\sentence_9$. The leader plan is illustrated with red arrows in the leader's view. The user sees only the follower's view during execution.}\label{fig:scenario}
\end{figure}

\citet{Suhr2019:cerealbar} describe \cerealbar in detail. 
\cerealbar is a two-player, turn-based game where a leader and follower collaborate to collect sets of matching cards. 
The game objective is to collect as many valid sets as possible in a 3D environment. 
The environment includes landmarks (e.g., houses, mountains, ponds, etc.) that the players must move around, and may obscure a player's view. 
A valid set consists of three cards with three distinct colors, shapes, and counts. 
Players move onto cards to select or deselect them. 
When the selected cards comprise a valid set, the players earn a point, all cards disappear,\footnote{In \citet{Suhr2019:cerealbar}, only the selected cards disappear. We introduced this modification to minimize inter-turn effects for the follower (i.e., memorize card locations).} and new cards appear. 
The two players must collaborate effectively using natural language.
The leader observes the entire environment, plans who should select which cards for the next set, executes their own part of this plan, and issues instructions to the follower.
The follower executes leader instructions, only seeing a partial first-person view of the environment. 
Leader instructions must make use of the observed spatial environment, including landmarks, for the follower to be able to execute them given their partial view. 
Each interaction includes multiple instructions. 
Figure~\ref{fig:scenario} shows the game and example generated instructions. 

\cerealbar was originally used for learning a follower instruction execution model from human demonstrations~\cite{Suhr2019:cerealbar}. 
In contrast, we learn an instruction generation model for the leader, with the human user as the follower. 
The generated instructions must often specify multiple tasks to complete (i.e., when the follower is to select multiple cards), and how to navigate to the target cards, because the follower has only partial observability of the environment. This includes references to landmarks, spatial relations, and descriptions of paths.
We focus on language generation, and use a deterministic planner to generate the plan, including which cards to select and how each player should move in their next turn, and execute the planned leader actions.
The system uses the model we learn to  map the follower's part of the plan to a natural language instruction.

We learn through interactions with non-expert human followers, which \cerealbar is particularly suited for. 
The utility-maximizing game objective to earn a high score by collecting as many valid sets as possible incentivizes followers to execute the generated instructions as accurately as possible. 
In addition, \cerealbar players need no expert knowledge to participate in the game, beyond familiarity with the simple game rules.

\begin{figure*}[t!]
    \centering
    \includegraphics[width=\textwidth]{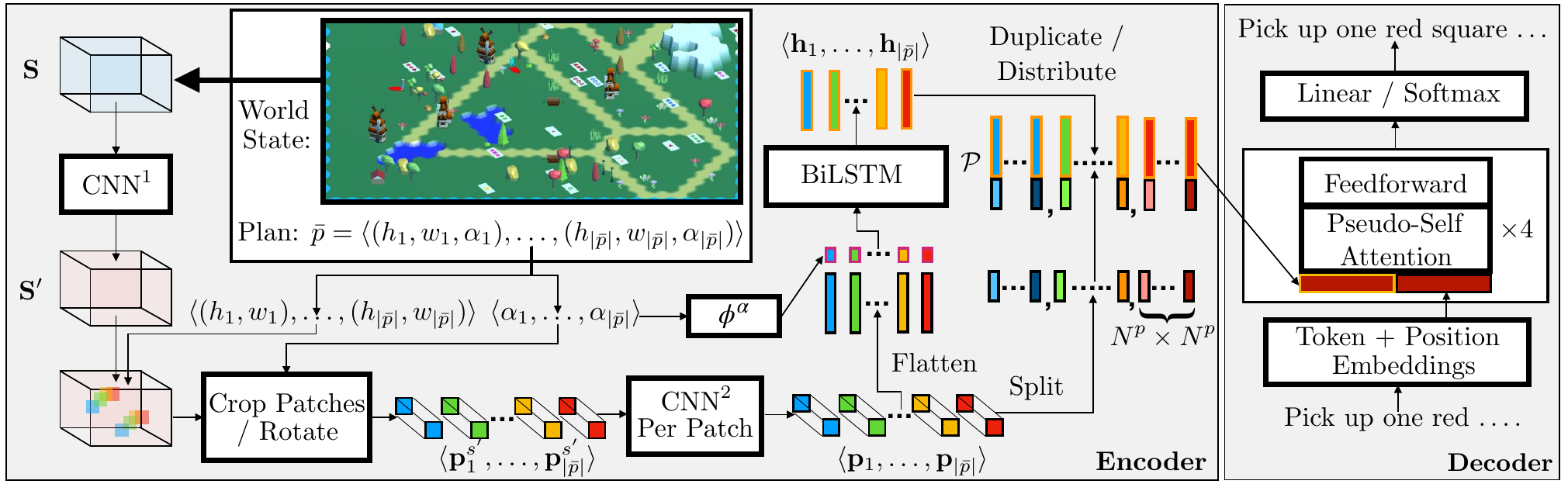}
    \caption{Model illustration. Section~\ref{sec:model} describes the model.}\label{fig:model}
\end{figure*}

\section{Model}\label{sec:model}

We design a relatively simple encoder-decoder architecture to model the generation distribution $\modelprob(\cdot \mid \startstate, \pathplan ; \params)$, leaving more complex model development for future work. The inputs are a start state $\startstate$ and a plan $\pathplan$. The model parameters are $\params$. 
Our design considers the environment and plan to generate relevant, grounded instructions. 
Figure~\ref{fig:model} illustrates the model. 

\paragraph{Inputs} 

Similar to \citet{Suhr2019:cerealbar}, we represent the world state $\startstate \in \{0,1\}^{P\times H\times W}$ as a binary 3D tensor, where $P$ is the number of position properties, and $H$ and $W$ are the environment's height and width. 
Each of the $W\times H$ positions is represented as a binary properties vector of length $P$ (e.g., encoding the type of object in the position, its color, etc.).
The system plan $\pathplan = \langle  \pose_1,\dots,\pose_{|\pathplan|}\rangle$ is a sequence of follower poses along the intended execution. Each pose $\pose_j$ is a tuple $(\heightidx_j, \widthidx_j, \rotidx_j)$ of height $\heightidx_j$ and width $\widthidx_j$ coordinates, and a discrete orientation angle $\rotidx_j$.

\paragraph{Encoder} 
The encoder computes a set of hidden states, which the decoder attends to during generation. 
We use a learned embedding function $\embedstate$ to map each position vector to a dense embedding of size $N^s$ by summing the embeddings of each of the position's properties.
We combine the embeddings into a tensor $\mathbf{S} \in \reals^{N^s\times H \times W}$, and compute: $\mathbf{S}' = \conv^1(\mathbf{S})$, where $\conv^1$ is a learned convolution and $\mathbf{S}' \in \mathbb{R}^{N^{s'} \times H \times W}$. Because the \cerealbar environment is a grid of hexagons, we use \textsc{HexaConv}~\cite{Hoogeboom:18hexaconv}. 
We encode the plan positions into a sequence of vectors $\langle \poserep_1^{s'}, \dots, \poserep_{|\pathplan|}^{s'} \rangle$ by cropping a $N^{s'} \times N^p\times N^p$-sized tensors from $\mathbf{S}'$ centered around each $(\heightidx_j, \widthidx_j)$ and rotated by $\rotidx_j$. 
These tensors represent the pose of the follower and its surroundings during execution. %
Each $\poserep_j^{s'}$ is encoded to $\poserep_j = \conv^2(\poserep_j^{s'})$, while retaining the dimensionality of $\poserep_j^{s'}$. 

We concatenate an orientation embedding $\embed^\rotidx(\alpha_j)$ to each $\poserep_j$, and process $[\poserep_1; \embed^\rotidx(\rotidx_1)], \dots,[\poserep_{|\pathplan|};\embed^\rotidx(\rotidx_{|\pathplan|})]$ with a bidirectional LSTM to compute $\mathbf{h}_1,\dots,\mathbf{h}_{|\pathplan|}$. 
We construct the set of hidden states $\mathcal{P}$ the decoder attends to by concatenating each $\mathbf{h}_j$ with the $N^p \times N^p$ position vectors encoded in each $\mathbf{p}_j$:

\begin{small}
    \begin{equation}    
        \begin{aligned}
            \mathcal{P} = \big\{ [\mathbf{h}_j; \poserep_{j}[x,y]] \mid  & 1 \leq j \leq  |\pathplan|, \\ & 1 \leq x,y \leq N^p \big\}\;\;,
        \end{aligned}
    \end{equation}
\end{small}

\noindent
where $\poserep_{j}[x,y]$ is a position vector of size $N^{s'}$. 

\paragraph{Decoder} 
The decoder computes a probability distribution over token types conditioned on the prefix generated so far and the set $\mathcal{P}$, which represents the environment state and plan.
The decoder uses the first four layers of the GPT-2 Transformer architecture~\cite{Radford2019:opengpt2}, which enables initializing with GPT-2 weights. 
We extend it with pseudo self attention~\citep{Ziegler:19psudoselfattention} to condition the generation on the encoder outputs $\mathcal{P}$. This adds a linear layer that projects the encoder outputs $\mathcal{P}$ into the decoder self-attention space. 

\paragraph{Inference}
We decode instructions from $\modelprob(\cdot \mid \startstate, \pathplan ; \params)$ using temperature sampling with a temperature of $\tau$~\cite{Kreutzer:18}. 
This sharpens the sampling distribution, to focus on higher probability outputs. 
We do not use beam search.

\section{Learning}\label{sec:learn}

We continually improve our model by observing users following generated instructions and re-estimating the model parameters. 
We initialize the model parameters $\params_1$ using an existing language model and training on a static dataset of instructions $\dataset_0$ (Section~\ref{sec:initialization}). 
We then perform a series of rounds, each round $r$ includes deploying the model with human users and training on the collected interactions (Section~\ref{sec:learn_from_users}).
In round $r$, we collect interactions between our model parameterized by $\params_r$ and human followers, to create a dataset $\dataset_r = \{(\startstate^{(i)}, \pathplanlabel^{(i)}, \sentence^{(i)}, \sentlabel^{(i)})\}_{i=1}^{|\dataset_r|}$ of start states $\startstate^{(i)}$, plans $\pathplanlabel^{(i)}$, instructions $\sentence^{(i)}$, and binary labels $\sentlabel^{(i)}$.
We estimate $\params_{r+1}$ using all data collected so far $\cup_{q=0}^r\dataset_q$.
Figure~\ref{fig:intro} illustrates our learning process.

\subsection{Initialization}\label{sec:initialization}

User interaction requires some level of minimal performance. 
Pilot experiments showed  that a poorly initialized system is likely to frustrate users, who in turn provide little learning signal. 
Our initialization provides a sufficient level of grammaticality and plausibility to support user interaction, and thereby further learning.

We initialize the decoder weights with the first four layers of GPT-2~\citep{Radford2019:opengpt2}. All other weights, including of the encoder and pseudo self-attention linear layers, are initialized randomly. 
We then train with a supervised dataset $\dataset_0 = \{(\state^{(i)}, \pathplanlabel^{(i)}, \sentence^{(i)}, \sentlabel^{(i)})\}_{i=1}^{|\dataset_0|}$ of human plans $\pathplanlabel^{(i)}$ starting at start states $\state^{(i)}$ and instructions $\sentence^{(i)}$, all with positive labels $\sentlabel^{(i)} = +1$. 
We use limited data, just sufficient to effectively interact with users for further learning. 
We estimate $\params_1$ by minimizing a supervised loss:

\begin{small}
\begin{equation}
    \begin{aligned}
    \mathcal{L}_I(\theta_1, &\dataset_0) =  \\  & - \frac{1}{|\dataset_0|} \sum_{i=1}^{|\dataset_0|} \log P (\sentence^{(i)} | \state^{(i)}, \pathplanlabel^{(i)}; \theta_1)\;.
    \end{aligned}
\end{equation}
\end{small}

\subsection{Learning from User Behavior}\label{sec:learn_from_users}

Learning from interacting with human users alternates between generating instructions in interaction with users and training the model.

\paragraph{Interaction with Users}

In each round $r$, we first deploy the model with parameters $\params_r$ to interact with human users, with our system as the leader and the user as the follower. 
We do not update the model during this interaction phase.

The game environment is randomly generated for each interaction. 
Each game continues until it concludes, either when the user leaves or the turns are exhausted.
A game often includes collecting multiple sets of cards, and generating multiple instructions. 
Each instruction is generated for the current state as the start state $\startstate$;\footnote{For simplicity, we do not index the game time step.} as both agents move and change the status of cards, the environment state changes throughout the game.
At state $\startstate$, we generate the plan $\pathplan$ using a deterministic planner that determines (a) which cards should be selected or de-selected to make the next valid set, and (b) the shortest paths the leader and follower should take to visit all target cards.
The actions the planner assigns to the follower form the plan $\pathplan$.
The actions assigned to the leader are executed by the leader agent deterministically during its turn.
The model is used to sample an instruction $\sentence \sim \modelprob(\cdot \mid \startstate, \pathplan; \params_r)$, which is displayed to the user.
The human user has no access to $\pathplan$, the set of target cards, or the game state $\startstate$. They only observe the instruction and what is ahead (Figure~\ref{fig:scenario}).

During their turn, the user executes $\sentence$ to the best of their ability, and indicates when done. 
If the user determines that the instruction cannot be followed, they can terminate the execution, which is treated just like marking the instruction as complete.  
The user execution $\execution$ is the entire sequence of poses they take while following the instruction. 

When the user concludes or terminates an instruction $\sentence$, we show them a top-down view of the entire environment with their execution path highlighted. They do not see the original system plan. 
We ask the user two binary feedback questions about the perceived correctness of their execution and grammaticality (Figure~\ref{fig:evalquestions}).

\begin{figure}[t]
    \centering
    \footnotesize
    \fbox{\begin{minipage}{0.98\linewidth}
        \textbf{Perceived correctness:} \nlstring{Did you follow all parts of the Leader's command and find everything correct?} \\
        Users are instructed to answer \nlstring{yes} only if they consider their execution correct given the instruction, if they perceive that the instruction describes the relevant cards and objects correctly, and if the specified actions make sense. \\[3pt]
        \textbf{Grammaticality:} \nlstring{Was the instruction grammatical and well written?} \\
        Users are shown examples of errors before the game, and largely interpret this criteria as language correctness independent of the world state. 
    \end{minipage}}
    \caption{The binary questions displayed to the user at the end of instruction execution.}\label{fig:evalquestions}
    \vspace{-5pt}
\end{figure}

We create a tuple $(\startstate, \pathplan, \sentence, \execution, \feedback)$ for each execution $\execution$, where $\startstate$ is the start state of the environment, $\pathplan$ is the plan generated in that state, $\sentence \sim \modelprob(\cdot \mid \startstate, \pathplan; \params_r)$ is the sampled instruction, and $\feedback$ is the set of responses to the feedback questions.
Once the user submits the answers to the feedback questions, the next instruction is generated.

\paragraph{Dataset Construction}

We use all interactions in round $r$ to construct dataset $\dataset_r$, which is made of tuples $(\startstate, \pathplanlabel, \sentence, \sentlabel)$, where $\pathplanlabel$ is a plan and $\sentlabel$ is a binary label.
Given a tuple $(\startstate, \pathplan, \sentence, \execution, \feedback)$, we use three heuristics to add examples to $\dataset_r$:
\begin{enumerate}[itemsep=2pt,topsep=0pt,parsep=0pt,partopsep=0pt]
    \item If any feedback answer in $\feedback$ is negative, the instruction does not reflect the user's execution or not well written (i.e., ungrammatical). We add a negative example to $\dataset_r$ with the system plan $\pathplan$: $(\startstate, \pathplan, \sentence, -1)$.
    \item If both feedback answers are positive, the user considers their execution $\execution$ accurate and the instruction well formed. This does not necessarily indicate the execution follows the system plan, but that we can treat the execution as a plan. We add a positive example with the execution as the plan: $(\startstate, \execution, \sentence, +1)$.
    \item If both answers are positive and the execution $\execution$ follows the plan $\pathplan$,\footnote{For instructions that target cards, we require getting the card selection right, and ignore the follower position. For instructions that require waiting (e.g., \nlstring{hold still}), we require the position to remain the same, but allow orientation deviation.} the instruction communicates the plan well. We add a positive example with the system plan: $(\startstate, \pathplan, \sentence, +1)$.
\end{enumerate}

Overall, we add examples to $\dataset_r$ using both the original system plan and the user execution. The heuristics utilize the observational learning signal as much as possible while avoiding examples not beneficial for learning. 
For example, we do not add negative examples using the user execution, because these are less likely to be useful for learning. Although such executions can form negative examples if the user answered negatively to the correctness question, they tend to be relatively arbitrary, and it is unlikely the model conditioned on them will assign significant probability to the generated instruction, which is the behavior negative examples come to suppress.

\paragraph{Parameter Estimation}
We estimate the model parameters for the next round $\params_{r+1}$ using all available data $\dataset = \cup_{q=0}^r\dataset_q$. 
We re-train our model, starting with GPT-2 parameters (Section~\ref{sec:initialization}).\footnote{Pilot studies showed re-training to be more stable than fine-tuning given new data, and we conduct the majority of our experiments with this method. However, we also observe that our process is overall robust to the initially observed instabilities of fine-tuning (Section~\ref{sec:results}).}

We formulate learning as an offline contextual bandit problem, treating the sentence labels $\sentlabel$ as rewards. 
Learning from the positive examples in $\dataset$ forms a straightforward supervised learning problem, albeit one where the data is generated from system interaction. 
A key challenge is using the negative examples. 
Treating them like supervised examples requires optimizing the probability of their instructions to zero. Because $\lim_{P(\cdot) \rightarrow 0}\log P(\cdot) = -\infty$, this leads to an unbounded negative loss that quickly dominates the objective. 
This in contrast to positive examples, for which the loss is bounded by zero. 
This issue is not present in existing work using offline contextual bandits to improve machine translation~\citep{Lawrence:17nmt,Kreutzer:18}, where rewards are always non-negative. 

We address this issue by adding an inverse propensity score~\cite[IPS;][]{Horvitz1952:ips,Wang2017:offpolicy-cb} coefficient to negative examples in a policy gradient objective. The gradient for estimating parameters $\params_{r+1}$ is:

\begin{small}
    \begin{equation}    
        \begin{aligned}
            &\nabla \mathcal{L}(\params_{r+1}, \dataset) =  \\ & \hspace{+2pt} \frac{1}{\dataset}\sum_{i=1}^{|\dataset|}\ipscoeff^{(i)}_{\params_{r+1}} \sentlabel^{(i)}\nabla\log \modelprob(\sentence^{(i)} \mid \state^{(i)}, \pathplanlabel^{(i)}; \params_{r+1})\;\;,
        \end{aligned}
    \end{equation}
\end{small}

\noindent
where, given an example $(\state^{(i)}, \pathplanlabel^{(i)}, \sentence^{(i)}, \sentlabel^{(i)})$ acquired in round $q$ with parameters $\params_q$, $\ipscoeff^{(i)}_\params$ is:

\begin{small}    
    \begin{equation}
        \ipscoeff^{(i)}_{\params} = 
            \begin{cases}
                1 & \sentlabel = +1\\
                \dfrac{\modelprob(\sentence^{(i)} \mid \state^{(i)}, \pathplan^{(i)}; \params )}{\modelprob(\sentence^{(i)} \mid \state^{(i)}, \pathplan^{(i)}; \params_q)}               & \sentlabel = -1 
            \end{cases}\;\;.
        \label{eq:score}
    \end{equation}
\end{small}

\noindent
As the probability of a negative example (i.e., $\sentlabel = -1$) decreases, so does its impact on the loss. 
While IPS is commonly used in bandit learning to de-bias the loss estimate~\cite{Lawrence:17nmt}, our motivation is different, and we do not add it to positive examples. Because of the large combinatorial space, sentence probabilities are generally small. The IPS coefficient of a positive example can become very large as its probability increases during learning. Instead, we use a supervised-like term, which is known to behave well.\footnote{An alternative, and important direction for future study is to add IPS to all examples, but clip it at a certain maximal value, similar to clipping in PPO~\cite{Schulman2017:ppo}.}

\section{Experimental Setup}\label{sec:exp}

\paragraph{Initialization Data}
We create the supervised initialization dataset $\dataset_0$ by sampling 360 interactions from the original \cerealbar data~\cite{Suhr2019:cerealbar}, which was collected in a wizard-of-oz~\cite[WOZ;][]{Kelley:84wizardofoz} setup via human-human games. 
We select this number through pilot studies and qualitative analysis to minimize the amount of initialization data, while still maintaining sufficient model performance for early interactions to facilitate learning. 
Our goal is to use as little data as possible to study the target scenario where investment in supervised data is minimal, and most learning is left to interaction with users. 
This data includes 7{,}147 examples. We use the human demonstrations in the original data as plans.

\paragraph{Evaluation}

Similar to \citet{Zhao21:eval-instruction-gen}, we observe that automated metrics, such as Bleu~\cite{Papineni02:bleu} or BERTScore~\cite{Zhang:20bertscore}, computed over a static held-out validation set are unreliable for evaluating instruction generation. 
Instead, we focus on task-completion measures via human execution.
We measure \textit{task completion} by considering the user execution as completing the intended task if the user visits all card locations included in the system plan; or, if the plan includes no target cards, the user stays in the starting position.
We quantify the similarity of the user execution to the path in the system plan by computing \textit{earth mover's distance}~\cite[EMD;][]{Rubner1998:emd}\footnote{We use POT~\cite{Flamary:21} to compute EMD.} between the two~\cite{Blukis2019:drone-sureal}. 
We also track the user answers to the feedback questions (Figure~\ref{fig:evalquestions}).
We average each measure over the number of instructions in each round. 

\paragraph{Language Analysis}

We quantitatively analyze how generated instructions change throughout training.
For each round, we report mean instruction length, vocabulary size, and three measures of syntactic complexity using dependency trees~\citep{Xu-Reitter2016:syntacticcomplexity}: (a) maximum depth: the longest path from root to a leaf; (b) maximum width: the maximum out-degree of any word in the tree; and (c) average branching factor: the average out-degree of non-leaf words. 
We normalize the three measures by instruction length. 
We qualitatively analyze errors in generated instructions, 
by comparatively analyzing 100 randomly sampled examples where the user failed to complete the intended task from the first and final rounds. 

\paragraph{Interaction Setup}

Except initialization, learning and evaluation are done through live interaction with users on Amazon MTurk. 
All workers passed a tutorial and a qualification quiz. 
We pay \$0.15 per interaction, with a bonus of \$0.10 per instruction to workers that follow our guidelines. 

\paragraph{Implementation Details}

Similar to performance evaluation, automated measures are unreliable for model selection.
Instead, for both initialization and in each round, we train for $N = 400$ epochs, and take the final model. 
We find $N$ via qualitative analysis of the initial model.
We use an ensemble of four models. 
We uniformly sample one of the four models to sample each instruction, and take its probability to use in IPS for negative examples. 
We use a sampling temperature $\tau = 0.5$, and AdamW~\citep{Loshchilov::18adamw} for learning.

\begin{figure*}[t!]
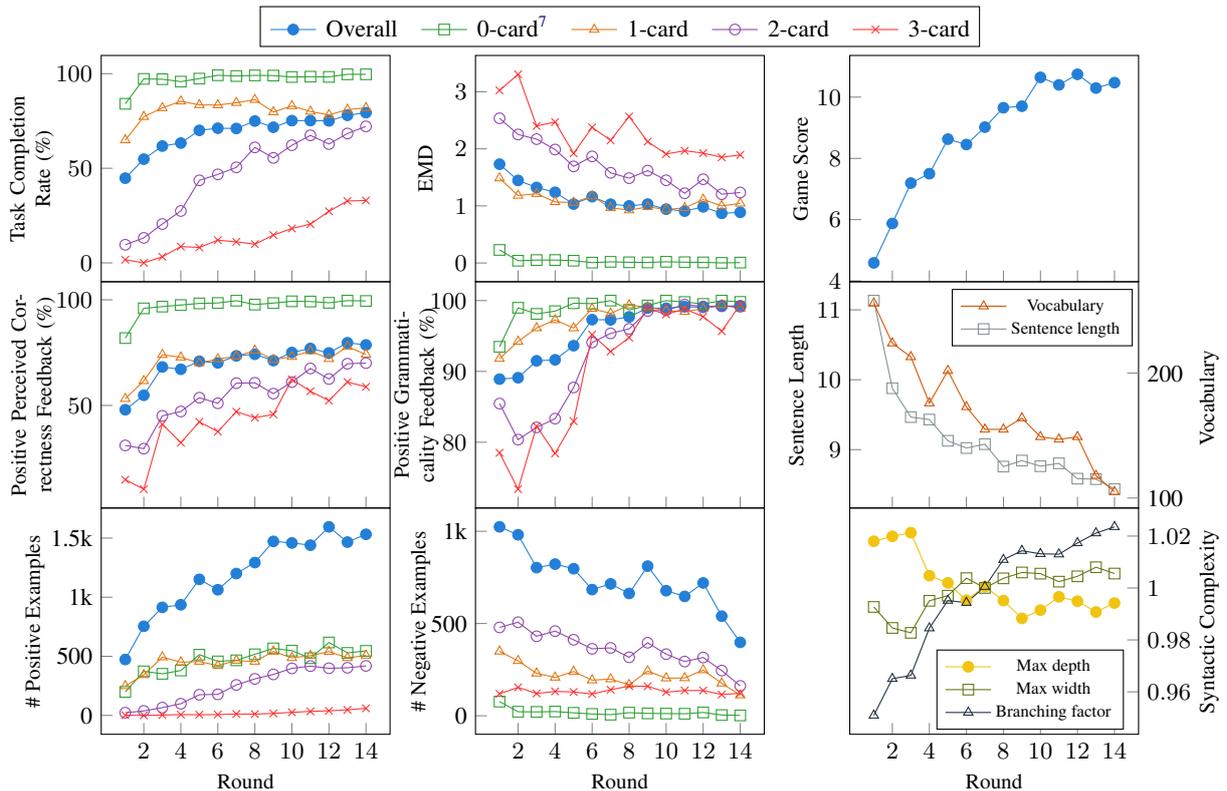

\footnotesize
\centering

\begin{tikzpicture}

\begin{groupplot}[
   group style={
      group size=3 by 3,
      vertical sep=0pt,
      horizontal sep=32pt,
      x descriptions at=edge bottom},
   width=3.8cm,
   height=3cm,
   ylabel={},
   ytick pos=left,
   ylabel shift = -3 pt,
   legend style={font=\scriptsize},
   xlabel style={font=\scriptsize},
   ylabel style={font=\scriptsize},
   xtick={2,4,6,8,10,12,14},
   xtick pos=bottom,
   scale only axis],
\nextgroupplot[ylabel style={align=center, text width=2cm},ylabel={Task Completion Rate (\%)}, ylabel shift = -5.5 pt]
\input{figures/main_results/tc}
\coordinate (top) at (rel axis cs:0,1); %
\nextgroupplot[ylabel={EMD}, ylabel shift = 2 pt]
\input{figures/main_results/emd}
\nextgroupplot[ylabel={Game Score}]
\input{figures/main_results/game_score}

\nextgroupplot[ylabel style={align=center, text width=3.5cm}, ylabel={Positive Perceived Correctness Feedback (\%)}, ylabel shift = -6 pt]
\input{figures/main_results/first_q}
\nextgroupplot[ylabel style={align=center, text width=4cm}, ylabel={Positive Grammaticality Feedback (\%)}, ylabel shift = -9.5 pt]
\input{figures/main_results/second_q}
\nextgroupplot[ylabel={Sentence Length}]
\input{figures/main_results/sentence_length}

\nextgroupplot[ylabel={\# Positive Examples}, scaled ticks=false, tick label style={/pgf/number format/fixed}, xlabel={Round}, yticklabels={0,0,500,1k,1.5k}, ylabel shift = -4 pt]
\input{figures/main_results/positive_examples}
\nextgroupplot[ylabel={\# Negative Examples}, scaled ticks=false, tick label style={/pgf/number format/fixed},
xlabel={Round}, yticklabels={0,0,500,1k}, ylabel shift = -6 pt]
\input{figures/main_results/negative_examples}
\nextgroupplot[ylabel={Syntactic Complexity}, xlabel={Round}, legend style={nodes={scale=0.8, transform shape}, at={(0.3,0.02)},anchor=south west},ytick pos=right]
\input{figures/main_results/dependency_trees}
\coordinate (bot) at (rel axis cs:1,0);%
\end{groupplot}

\path(top|-current bounding box.north)--
  coordinate(legendpos)
  (bot|-current bounding box.north);
\matrix[
    matrix of nodes,
    anchor=south,
    draw,
    inner sep=0.2em,
    draw
  ]at([yshift=3pt]legendpos)
  {
    \ref{plots:oveall}& Overall&[5pt]
    \ref{plots:0-card}& 0-card\textsuperscript{\ref{fn:zerocard}}&[5pt]
    \ref{plots:1-card}& 1-card&[5pt]
    \ref{plots:2-card}& 2-card&[5pt]
    \ref{plots:3-card}& 3-card\\};

\begin{groupplot}[
   group style={
      group size=3 by 3,
      vertical sep=1pt,
      horizontal sep=32pt,
      x descriptions at=edge bottom,
      y descriptions at=edge right,
      },
   width=3.8cm,
   height=3cm,
   ylabel={},
   xtick=\empty, axis line style=transparent,
   ytick={100,200,300},
   yticklabels ={},
   legend style={nodes={scale=0.8, transform shape},font=\scriptsize},
   xlabel style={font=\scriptsize},
   ylabel style={font=\scriptsize},
   xtick style={draw=none},
   xticklabels=\empty,
   scale only axis],
\nextgroupplot[ytick style={draw=none}]
\nextgroupplot[ytick style={draw=none}]
\nextgroupplot[ytick style={draw=none}]
\nextgroupplot[ytick style={draw=none}]
\nextgroupplot[ytick style={draw=none}]
\nextgroupplot[ylabel={Vocabulary}, ytick={100,200,300}, yticklabels={100,200,300}, ytick pos=right]
\input{figures/main_results/vocabulary}
\addlegendimage{plotcolor9, mark=square}
\addlegendentry{Vocabulary}
\addlegendimage{plotcolor10, mark=triangle}
\addlegendentry{Sentence length}
\nextgroupplot[ytick style={draw=none}]
\nextgroupplot[ytick style={draw=none}]
\nextgroupplot[ytick style={draw=none}]
\end{groupplot}
\end{tikzpicture}
\caption{The system's lifetime statistics from the long-term experiment (14 rounds). 
The system improves on task completion ($\uparrow$), EMD ($\downarrow$), positive response rate for the two feedback questions ($\uparrow$), and  game score ($\uparrow$). 
Section~\ref{sec:results:longterm} discusses these results in detail.
}\label{fig:fullexp}
\end{figure*}

\section{Results and Analysis}\label{sec:results}

We conduct a long-term experiment with 14 rounds using our approach, and separate seven-round experiments to compare system variants. 
In both experiments, we collect roughly 100 interactions for each system per round. 
In the seven-round experiments, we deploy methods simultaneously to ensure that our observations are not sensitive to changes in user behavior, for example because of adaptation and increased expertise. 
We do not inform workers about the model they are interacting with.
We train each system only on data collected by the same method in previous rounds. 

\subsection{Long-term Study}\label{sec:results:longterm}

We experiment with our approach for 14 rounds. 
We collect a total of 27{,}031 instructions from 1{,}445 interactions, with 103.2 interactions per round on average. The total cost is \$2{,}895. 
Figure~\ref{fig:fullexp} shows both performance measures and language trends. 
For task measures and user feedback, we also break down performance according to the number of target cards in the system plan to evaluate performance changes for plans which may be more difficult to describe (e.g., because they require specifying more cards).\footnote{0-card plans target no cards (e.g., \nlstring{hold still}).\label{fn:zerocard}}

Our learning method significantly improves the system performance across all measures. 
Task completion rate improves from 44.7\% at round one to 79.3\% at round 14, while EMD decreases from 1.73 to 0.88, showing increasing similarity between the execution and the plan. 
The user perception of the system also improves: the positive response rate for the perceived correctness question improves from 47.9\% to 78.6\%, and for grammaticality from 88.9\% to 99.2\%. 
The overall collaborative system performance improves as well; the game score increases from 4.5 to 10.4. 
The number of positive examples per round gradually increases, as the system improves and the interactions become longer. In contrast, the number of negative examples decreases over time.

We observe that the initial model struggles to describe plans containing more target cards, with a particularly low task completion rate of 1.6\% for 3-card plans in the first round. 
This is potentially because only 0.7\% of human follower executions in $\dataset_0$ demonstrate picking up three cards, while the planner generates 3-card plans 7.9\% of the time.
While initial improvement is slow for 3-card instructions, it picks up around round eight, and reaches 32.9\% task completion rate. 

\paragraph{Language Analysis}
We observe a consistent trend of decreasing sentence length and vocabulary size.
Overall, these trends accompany reduction in over generation of erroneous phrases that are not grounded well in the environment. 
We also qualitatively observe that the systems gradually generates slightly more underspecified instructions, for example by dropping mentions of landmarks  crucial for navigating to a target card.
This may explain the slight decrease in 1-card task completion rate in later rounds (Figure~\ref{fig:fullexp}), because the planner usually has the follower travel further for 1-card instructions, which requires relying more on landmarks. 
A potential explanation to the decrease in vocabulary size is the ever increasing presence of system-generated sentences in training, which reinforces the system's word choices. 
Alternatively, our learning signal may not account for the need for more descriptive language.  For example, humans may compensate with exploration for omitted descriptions, which is not distinguished by how we convert the observed  behavior to a learning signal. 
These trends outline important directions for future work. 

\newcommand{\removestuff}[1]{}

\begin{table*}[t!]
    \footnotesize\centering
    \addtolength{\tabcolsep}{-2pt}
    \begin{tabular}{p{4.9cm}ccp{7.1cm}}
        \toprule
        \textbf{Error Type} & $r=1$ & $r=14$ &
        \textbf{Example} \\
        \midrule
        Incorrect, missing, or extra cards  & \removestuff{51.4\%} 75 & \removestuff{10.5\%} 39 & \nlstring{turn left and go to the yellow \erase{star}  \textcorrection{triangles}}\\
        Irrelevant landmarks & \removestuff{\phantom{0}8.9\%} 13  & \removestuff{\phantom{0}0.3\%} \phantom{0}1  & \nlstring{Head toward the \erase{windmill} \textcorrection{house}. grab 2 red and triangle}\\
        Incorrect direction & \removestuff{20.6\%} 30 & \removestuff{\phantom{0}9.4\%} 35 & \nlstring{grab the black heart to \erase{your left} \textcorrection{in front of you}.} \\
        Incorrect actions or conditions &  \removestuff{19.2\%} 28 & \removestuff{\phantom{0}3.8\%} 14 & \nlstring{\erase{After the two red triangles}, get the 3 red triangles.} \\
        Underspecification & \removestuff{\phantom{0}5.5\%} \phantom{0}8 & \removestuff{\phantom{0}7.0\%} 26 & 
        \nlstring{\textcorrection{turn right and go straight toward red trees} collect two orange triangle.} \\
        Implausible instructions & \removestuff{\phantom{0}7.5\%} 11  & \removestuff{\phantom{0}0.3\%} \phantom{0}1   & \nlstring{Turn left and get the two pink hearts \erase{and the two pink hearts near the pink hearts.}} \\
        \midrule
        \textbf{Proportion of erroneous instructions}  & 68.5\%  &  26.8\%   & \\
        \bottomrule
    \end{tabular}
    \caption{
    The types of errors observed in erroneous instructions generated during the first ($r=1$) and final ($r=14$) rounds of deployment.
    We show error counts from the 100 randomly-sampled erroneous instructions. 
    Examples illustrate error categories; \textred{red} strikethrough shows erroneous segments, and \textcorrection{blue} fragments show possible corrections.
    Instructions that fit into multiple categories are double counted.}\label{tab:errors}
\end{table*}

We observe a small increase in syntactic complexity over the system's lifetime with regard to the branching factor, which shows significant increase ($p < 0.00001$).\footnote{We use $t$-test ($\alpha=0.01$) comparing rounds 1 and 14.} 
We also see a slight decrease in maximum tree depth ($p < 0.0001$), and no significant change in max width.

\paragraph{Error Analysis} 

We analyze errors in the generated instructions at the first and final rounds.
For each round, we randomly sample 100 instructions that the user did not execute according to the plan or answered negatively to a feedback question. 
Table~\ref{tab:errors} shows error types and example instructions. 
Overall, the frequency of erroneous instructions decreases from 68.5\% of instructions in the first round, to 26.8\% in the final round. 
From the first to final round, we observe noticeable decrease in errors related to grounding of cards and landmarks.
The overall frequency of errors related to incorrect directions and incorrect actions or conditions also decreases, and implausible instructions diminish close to zero percent.
However, there is an overall increase in underspecified instructions. This aligns with the decrease in the vocabulary size and landmark use we discuss above.

\begin{table}[t] 
    \footnotesize\centering
    \addtolength{\tabcolsep}{-3pt}
    \begin{tabular}{lllllll}
        \toprule
        \textbf{Model} & $r$ & \textbf{Overall} & \textbf{0-card}\textsuperscript{\ref{fn:zerocard}} & \textbf{1-card} & \textbf{2-card} & \textbf{3-card} \\
        \midrule 
         $\params_1$ & 1 & 44.8 & 84.1 & 64.9 & \phantom{0}9.6 & \phantom{0}1.7\\
         $\params_1$ & 14 & 45.1 & 84.5 & 62.1 & \phantom{0}9.3 & \phantom{0}0.8\\
         $\params'_1$ & 14 & 49.6 & 76.6 & 63.8 & 24.8 & \phantom{0}7.4\\
         $\params_{14}$ & 14 & \textbf{79.4} & \textbf{99.6} & \textbf{81.9} & \textbf{72.1} & \textbf{33.0}\\
        \bottomrule
    \end{tabular}
    \caption{
    The effect of confounding factors on task completion rate (\%).
    The initial model $\params_1$ is evaluated both in the first ($r=1$) and final ($r=14$) rounds, showing no effect of user adaptation. 
    In the final round, we also evaluate $\params'_1$, which is trained on the same data as $\params_1$ but using more gradient updates.
    We also show results for the final-round model $\params_{14}$. 
    }\label{tab:confounding_factors}
\end{table}

\paragraph{Confounding Factors} 
We identify two mechanisms unrelated to our approach that could explain the observed performance changes. 
We deploy two additional systems alongside our system during the final round. 
For each interaction, one of the three systems is randomly chosen. We do not inform the workers of the identity of the model for each interaction.
First, we deploy the system following initialization during the final round to study if performance might be explained by user improvement over time. 
Second, because we train with a fixed number of epochs, later rounds have many more gradient updates, which may allow for better parameter estimation, even with the same amount of data. 
We train a system on the initialization dataset $\dataset_0$ for the same number of gradient updates as when training the final full system. 

Table~\ref{tab:confounding_factors} shows these confounding factors do not explain the observed gains.  
We find minimal differences between evaluating the initial model ($\params_1$) at the beginning and end of deployment, showing no significant effect from user improvement. 
Training the initial system longer ($\params'_1$) shows a slight overall improvement, but negligent compared to final system ($\params_{14}$). 

\subsection{System Variants Study}\label{sec:results:variants}

We vary different design decisions, and experiment for seven interaction rounds.\footnote{This study is similar to ablation analysis, but aims to study different learning design decisions. Full-fledged repetitive ablations to identify the ideal system design are particularly challenging in this work, both because of experiment costs and the complex dynamics of interacting with users.} 
We experiment with four system variants:
(a) \sysfull: our full approach described in Section~\ref{sec:learn}; 
(b) \sysposonly: use only examples with positive labels $\sentlabel=+1$; 
(c) \systconly: ignore the feedback questions, instead if the user completes the task according to our task success measure we add positive examples with both the system plan and user execution, otherwise we add a negative example using the system plan; 
(d) \sysnoensemble: train and deploy a single model each round, starting from an initial model randomly sampled from these we use for \sysfull; 
and (e) \sysfinetune: train model parameters $\theta_{r+1}$ on $\dataset_r$ for $N$ epochs, starting from $\theta_r$, avoiding overfitting with rehearsal~\citep{Rebuffi2017:icarl,Hawkins::20continualcommunication}. In rehearsal, in each batch, half the examples are sampled randomly from the previous datasets $\dataset_0$,\dots,$\dataset_{r-1}$.
Except the variations specified, the systems are identical. 
We do not deploy a system ablating IPS, because we observe that training with negative examples without IPS results in a largely unusable system. 

\newcommand{\sysvarmarkfull}{*}
\newcommand{\sysvarmarkposonly}{square}
\newcommand{\sysvarmarktconly}{triangle*}
\newcommand{\sysvarmarknoens}{o}
\newcommand{\sysvarmarkfinetune}{diamond}

\begin{figure}
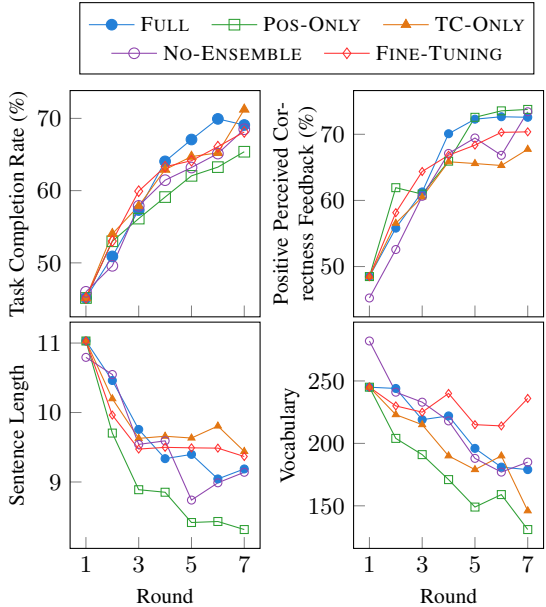

    \centering
    \footnotesize
    \begin{tikzpicture}
    
    \begin{groupplot}[
      group style={
        group size= 2 by 2,
        vertical sep=2,
        horizontal sep=35pt,
        x descriptions at=edge bottom},
      width=2.5cm,
      height=3cm,
      xlabel style={font=\scriptsize},
      ylabel style={font=\scriptsize},
      ylabel={},
      xlabel={Round},
      ytick pos=left,
      ylabel shift = -3 pt,
      xtick={1,3,5,7},
      xtick pos=bottom,
      scale only axis,
      ],
    \nextgroupplot[ylabel={Task Completion Rate (\%)}, mark size=2pt]
    \coordinate (top) at (rel axis cs:0,1); %
    \input{figures/system_comp/tc}
    \nextgroupplot[ylabel style={align=center, text width=3.5cm}, ylabel={Positive Perceived Correctness Feedback (\%)}, ylabel shift = -5 pt, mark size=1.5pt,
    ]
    \input{figures/system_comp/first_q}
    \nextgroupplot[ylabel={Sentence Length}, mark size=1.5pt,
    ]
    \input{figures/system_comp/sentence_length}

    \nextgroupplot[ylabel={Vocabulary}, ylabel shift = -3 pt, mark size=1.5pt,
    ]
    \input{figures/system_comp/unique_vocab}

    \coordinate (bot) at (rel axis cs: 1,0); %
    \end{groupplot}
    
    \path(top|-current bounding box.north) --
      coordinate(legendpos)
      (bot|-current bounding box.north);

    \node[
        matrix of nodes,
        anchor=south,
        draw,
        inner xsep = 0.2em,
        node font=\scriptsize,
        draw
      ]at([yshift=-1ex, xshift=0ex]legendpos)
      {
        \ref{plots:ensemble} \sysfull \hspace{0.5em}
        \ref{plots:pos_only} \sysposonly \hspace{0.5em}
        \ref{plots:tc_only} \systconly & \\
        \ref{plots:no_ensemble} \sysnoensemble \hspace{0.5em}
        \ref{plots:fine_tuning} \sysfinetune\\};
    \end{tikzpicture}
    \caption{Comparison of system variants.}
    \label{fig:short_sys_comparisons}
\end{figure}

We collect a total of 63{,}189 instructions across all systems, with 3173 interactions. Each round includes 453.2 interactions on average. The total cost is \$7{,}165. 
All systems are used concurrently in each round, including re-deploying \sysfull again starting from initialization. 
Figure~\ref{fig:short_sys_comparisons} shows the results. 
Despite some differences between the system variants, our method is largely robust to variations in learning design decisions.

All systems achieve comparable improvements in task completion rate, except for \sysposonly, which slightly underperforms.
We observe faster decrease in the vocabulary size and instruction length for \sysposonly, which does not use negative examples.  
This is possibly because the loss from negative examples encourages a more uniform generation distribution, potentially slowing down the overall trends of making the generation distribution more peaky. 
\systconly, which ignores the answers to user feedback questions when constructing the dataset, shows fewer positive responses to the perceived correctness feedback, although task completion remains comparable. 

We observe that using a single (\sysnoensemble) model rather than an ensemble leads to limited difference in overall performance. 
However, because of the challenge of identifying a good automated metric to stop training, the performance of models following training varies significantly. 
This can lead to deploying a bad model, which provides users with a poor experience. 
Using an ensemble of models incurs higher computational cost, but makes such a worst-case scenario less likely. 
For example, in our long-term experiment, the maximum task completion performance gap we observe between the best and worst models in each round is 13\%. 

Finally, we observe that fine-tuning (\sysfinetune) works as well as our re-training approach (\sysfull), potentially with a more stable vocabulary size. 
This is in contrast to our initial experiments, which showed it is harder to get consistent improvements through fine-tuning. While the fine-tuning process is harder to design  because it requires to choose the fine-tuning procedure (e.g., rehearsal~\citep{robins:95forgetting} or  KL regularization~\citep{yu13:klreg}) and carefully optimize additional hyperparameters, it can work just as well as re-training. Because fine-tuning is faster to train between rounds, it may be preferable in future work. 

\subsection{Comparison to Supervised Learning}

We also separately study the learning trends of our method compared to training on equivalent amount of supervised WOZ data. 
Supervised data is fundamentally different from our bandit data, for two main reasons: (a) it is significantly costlier because it requires a dedicated instruction-writing effort, whereas our data arises naturally from the system interaction with users during deployment; and (b) it provides per-token labels, whereas our data includes only utterance-level binary labels.
For the supervised system, after each round, we expand the dataset by randomly drawing an equivalent amount of additional data from the complete dataset of \citet{Suhr2019:cerealbar}, which includes 19{,}112 examples from 960 interactions.\footnote{Interactions with the supervised system are not used for learning, but only for evaluation.} 
This dataset allows for seven rounds. 
We  concurrently deploy a no-ensemble variant of our continual learning system.
We collect a total of 22{,}216 instructions across both systems, with 1{,}166 interactions. This experiment's total cost is \$2{,}230.

\begin{figure}[t]
    \centering
    \footnotesize
    \begin{tikzpicture}
    
    \begin{groupplot}[
      group style={
        group size = 1 by 1,
        vertical sep=0pt,
        horizontal sep=35pt,
        x descriptions at=edge bottom},
      width=5cm,
      height=3cm,
      xlabel style={font=\scriptsize},
      ylabel style={font=\scriptsize},
      ylabel={},
      xlabel={Round},
      ytick pos=left,
      ylabel shift = -3 pt,
      ytick={0,25,50,75,100},
      xtick={1,2,3,4,5,6,7,8,9},
      xtick pos=bottom,
      scale only axis,
      ],

    \nextgroupplot[ymin=-5, ymax=105, ylabel style={align=center, text width=3.5cm},
    ylabel={Task Completion Rate (\%)}, ylabel shift = -5 pt, mark size=1.5pt]
    \input{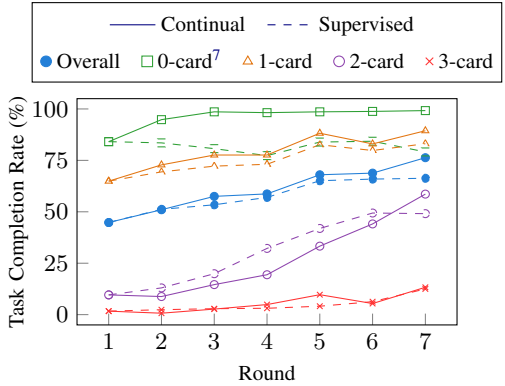}

    \iffalse
    \nextgroupplot[ymin=40, ymax=80, ylabel={Overall Task Completion Rate (\%)}, mark size=2pt]
    \coordinate (top) at (rel axis cs:0,1); %
    \input{figures/supervised_data/overall_tc}
    \iffal
    \nextgroupplot[ymin=-5, ymax=105, ylabel style={align=center, text width=3.5cm}, ylabel={Other Task Completion Rate}, ylabel shift = -5 pt, mark size=1.5pt,
    ]
    \input{figures/supervised_data/other_tc}
    \fi
    
    \coordinate (bot) at (rel axis cs: 1,0); %
    \end{groupplot}
    
    \path(top|-current bounding box.north) --
      coordinate(legendpos)
      (bot|-current bounding box.north);

    \node[
        matrix of nodes,
        anchor=south,
        draw,
        inner xsep = 0.2em,
        node font=\scriptsize,
        align=center,
      ]at([yshift=-1ex, xshift=0ex]legendpos)
      {
        \hspace{1.5em}\ref{plots:ours-line-symbol} Continual \hspace{0.5em}
        \ref{plots:sl-line-symbol} Supervised
        \hspace{1.5em}
        \\
        \ref{plots:overall-symbol} Overall  \hspace{0.5em}
        \ref{plots:0-card-symbol} 0-card\textsuperscript{\ref{fn:zerocard}} \hspace{0.5em}
        \ref{plots:1-card-symbol} 1-card 
        \hspace{0.5em}
        \ref{plots:2-card-symbol} 2-card \hspace{0.5em}
        \ref{plots:3-card-symbol} 3-card\\};
    \end{tikzpicture}
    \caption{Comparison to supervised learning. The continual learning system is competitive in task completion rates with systems trained on equivalent amount of supervised data. 
    }
    \label{fig:supervised_comparison}
\end{figure}

Figure~\ref{fig:supervised_comparison} shows our continual learning system consistently outperforms this supervised alternative in  overall task completion rate. 
There are two potential explanations to this gap. First, the data our approach uses is made of examples the system is likely to generate, potentially providing a more effective learning signal. Second, there is a difference between the plans of human leaders and our planner. Our training is better suited to adapt to how the complete system is designed, whereas training on human-annotated data is bound to suffer from a distribution shift. 
However, the continual learning system did not consistently outperform the supervised alternative on 2-card and 3-card instructions, especially at early rounds.
This is likely because the continual learning system generates few positive examples for more complex system plans (i.e., 2-card or 3-card) at earlier rounds. At later rounds, as the system improves, we observe more positive examples for such plans, creating an accelerating effect of improvement, which is best observed in our long-term experiment (Figure~\ref{fig:fullexp}).

\section{Related Work}\label{sec:related}

Learning for instruction generation has been studied using supervised methods, with examples of task specifications (i.e., contexts) paired with human-written  instructions~\cite[e.g.,][]{Daniele:16,Narayan-Chen2019:minecraft-dialogue}, including to improve instruction following~\cite{Fried:17pragmatic-models,Tan:19backtranslation}. 
We focus on continually learning by observing users executing generated instructions. 
This reduces annotation needs, and delegates much of the learning to interaction with users during system deployment. 
Language generation in context was also studied in scenarios that are not explicitly instructional, but aim to elicit specific behavior, such as negotiation games~\cite[e.g.,][]{Lewis:17negotiation} and referring expression generation~\cite[e.g.,][]{Dale1995:generation-gricean}. 

\citet{Gatt2017:nlg-survey} survey existing work on language generation, including using rule-based methods. 
Similar to our approach, some rule-based methods were evaluated with human followers in situated environments using task success~\cite[e.g.,][]{Koller2010:give2-nlg-report,Janarthanam2011:gruve-nlg-challenge}. 
Such methods are accurate and reliable, but are limited to pre-specified rules and remain static following development. 
Our focus is on studying the potential for learning by observing human behavior. The two approaches can be combined, for example by using rule-based methods to generate initialization data for our approach. 

Bandit learning has been studied with simulated user ratings for machine translation~\cite{Nguyen2017:bandit-mt,Lawrence:17nmt,Kreutzer:17} and semantic parsing~\cite{Lawrence:18}. 
We learn from real users, similar to recent studies in machine translation~\cite{Kreutzer:18naacl,Kreutzer:18}. 
In general, such learning assumes users can judge the system output, for example via proficiency in the language they wish to translate to. 
Our learning signal does not require such expertise, and is available naturally from the interaction. 

Explicit human feedback has also been incorporated into reinforcement learning methods~\cite{Knox:09interactiveshaping,Pilarski:11,Daniel:15,Mathewson:16,Warnell:17deep-tamer,MacGlashan:17humanfeedback,Arumugam2018:deep-coach}, including in the context of dialogue system learning~\cite{Liu2018:dialogue-e2e-human-feedback}.
\citet{Jacques:20} study forming a reward from implicit feedback for non-task-oriented dialogue language generation, by training multiple models to detect linguistic signals, such as sentiment and lexical overlap, that correlate with explicit user feedback. 
Learning from users has also been studied by asking users to rank system outputs~\cite[e.g.,][]{Wilson:12,Christiano:17deeprl_humans}, including for instruction following~\cite{Wang:16games} and summarization~\cite{Stiennon2020:summarization-feedback}. 
Unlike our approach, such ranking requires knowing the true system intent, and is not part of the system's normal operation (i.e., instructing users in our case).

Incorporating human users into learning is related to active learning~\cite{Settles2009:active-learning}, where a policy selects examples for an oracle to label during learning. 
Unlike common active learning scenarios we do not select examples from a static underlying distribution (i.e., a training set) for annotation, but generate examples with the learned model. 
This is similar to query synthesis active learning~\cite{Angluin1988:active-learning}, where examples are generated for annotation, rather than being selected from a set of unannotated examples. 
A more significant difference is that active learning methods solicit model output annotations by presenting an oracle with model inputs. 
In contrast, our approach exposes users to model outputs (i.e., generated instructions). It does not solicit written instructions, as would be expected if requesting labels. We also do not show model inputs (i.e., plans) to users. 
Finally, our model interacts with users during system operation, while completing its task. It does not require oracle  annotators. 

Language learning from behavioral signals has been studied in the cognitive science and psychology literature.\footnote{This review is not comprehensive, and only aims to highlight the relation to problems studied in related disciplines.} 
\citet{Krauss1966:concurrent-feedback-confirmation} study two types of feedback in human studies: concurrent linguistic feedback and behavioral intent confirmation, and show how both influence linguistic adaptation in an interaction over time. 
Studies of reference games reproduced the effect of confirmation feedback, showing that successful intent communication reinforces convention formation in the form of shorter references~\cite{Clark1986:colab-referring,Hawkins2020:learning-dynamics-ref-games}. 
Our learning signal is a type of confirmation feedback. 
However, our interaction procures and makes use of more complex learning signals than a simple binary intent communication success, by using the path the listener takes in response to the generated instruction as an alternative intent when constructing data for learning (Section~\ref{sec:learn_from_users}).\footnote{In more recent reference games~\cite{Hawkins2020:learning-dynamics-ref-games}, unlike in \citet{Krauss1966:concurrent-feedback-confirmation}, the choice of a bad referent can be seen as related to our use of listener execution.}

\section{Discussion}\label{sec:disc}

We propose a methodology to continually improve an instruction generation model by observing human users executing natural language instructions, and demonstrate its efficacy within a collaborative instruction following scenario. 
Our study shows that observation of user behavior is an informative signal for generating language to relay instructional intent. 
To the best of our knowledge, this type of learning signal has not been studied before.
This learning setting facilitates continual learning through interaction with users, and is particularly compelling for interactions with collaborative agents, including robots and software agents. 
Such agents are likely to operate in constantly changing environments (e.g., robots in homes), where continual learning is necessary to adjust to changes. 
Our continual learning approach also provides systems the flexibility to co-adapt to human users, who are likely to change preferences and behaviors in response to system behavior. 

Our experiments demonstrate the learning process is robust to various learning and process design choices. 
However, they also show it is accompanied by a reduction of language complexity, including reducing the effective vocabulary and sentence length. 
While much of the decrease in the effective vocabulary size throughout the system lifetime relates to generating fewer erroneous phrases, it also reduces the language diversity and descriptiveness. 
Our experiments show that this trend can be slowed down by using negative examples, and appears to be less pronounced when using fine-tuning. 
The combination of this decrease with the preference for shorter instructions makes it difficult for the system to describe longer, complex trajectories. 
Qualitatively, we observe this open problem is responsible for a significant portion of the remaining errors.
An important direction for future work is experimenting with directly encouraging more diverse language. 
This can be combined with approaches that allow for introducing new word types, which is unlikely in our approach, even though it uses sub-word tokenization. 
A potential direction in this vein is combining active learning to solicit human-written oracle instructions for plans the system fails to communicate. 

Our work highlights several other directions for future work. 
There is a strong need for a reliable automated metric to evaluate instruction generation. 
In absence of such a metric, we use a simple, but likely sub-optimal stopping criteria for learning. 
Beyond the learning signal we explored in our experiments, there are additional potential cues available during interaction.
For example, using continuous-valued similarity between system intent and user execution, modeling follower quality to discount the learning signal from interactions with bad followers, or weighing the feedback questions differently for more nuanced reward. 

Finally, the decrease in utterance length and vocabulary size mirrors similar trends observed in studies of human communication~\cite{Clark1986:colab-referring,Hawkins2020:learning-dynamics-ref-games}. 
This illustrates the potential of continual learning systems to reflect the dynamics of language change human participants expect in natural language interactions.
Observations of human learning also indicate the potential of integrating our approach with conversational  self-repair~\cite{Clark2020:conversational-repair} and partner reformulation~\cite{Clark2018:conversation-lang-acquisition}, both important components of child language acquisition that likely provide better credit assignment for learning compared to our binary bandit signal.

\section*{Acknowledgments}

This research was supported by ARO W911NF-21-1-0106, a Google Focused Award, Masason Foundation, a Facebook Fellowship, and NSF under grants No. 1750499 and DGE-1650441. 
We thank Jonathan Chang, Sasha Rush, the Cornell NLP Group, Robert Hawkins, Dipendra Misra, and John Langford for discussion and comments; Suyi Diao for Unity development; Anna Effenberger for code to compute syntax complexity; Ge Gao, Koji Shiono, and Takayuki Kojima for feedback on our interaction platform; and the crowdsourcing workers for participating in our data collection.   
Finally, we thank the action editor and the anonymous reviewers for detailed comments.

\bibliography{local,main}
\bibliographystyle{acl_natbib}

\end{document}